\pdfoutput=1

\documentclass[11pt]{article}

\usepackage[]{EMNLP2022}

\usepackage{times}
\usepackage{latexsym}
\usepackage{graphicx}
\usepackage{amsmath}
\usepackage{amssymb}
\usepackage{booktabs}
\usepackage{multirow}

\usepackage{xspace}
\usepackage[T1]{fontenc}

\usepackage[utf8]{inputenc}
\newcommand{\nop}[1]{}
\usepackage{listings}

\usepackage{microtype}

\usepackage{inconsolata}
\newcommand{\modelname}{MuRAG\xspace}

\title{MuRAG: Multimodal Retrieval-Augmented Generator \\ for Open Question Answering over Images and Text}

\author{Wenhu Chen, Hexiang Hu, Xi Chen, Pat Verga, William W. Cohen\\
  Google Research \\
  \small \texttt{\{wenhuchen,hexiang,patverga,wcohen\}@google.com}
  \\}

\begin{document}
\maketitle
\begin{abstract}
While language Models store a massive amount of world knowledge implicitly in their parameters, even very large models often fail to encode information about rare entities and events, while incurring huge computational costs. Recently, retrieval-augmented models, such as REALM, RAG, and RETRO, have incorporated world knowledge into language generation by leveraging an external non-parametric index and have demonstrated impressive performance with constrained model sizes. However, these methods are restricted to retrieving only textual knowledge, neglecting the ubiquitous amount of knowledge in other modalities like images -- much of which contains information not covered by any text. To address this limitation, we propose the first Multimodal Retrieval-Augmented Transformer (MuRAG), which accesses an external non-parametric multimodal memory to augment language generation. MuRAG is pre-trained with a mixture of large-scale image-text and text-only corpora using a joint contrastive and generative loss. We perform experiments on two different datasets that require retrieving and reasoning over both images and text to answer a given query: WebQA, and MultimodalQA. Our results show that MuRAG achieves state-of-the-art accuracy, outperforming existing models by 10-20\% absolute on both datasets and under both distractor and full-wiki settings.
\end{abstract}

\section{Introduction}
\begin{figure}[!thb]
    \centering
    \includegraphics[width=1.0\linewidth]{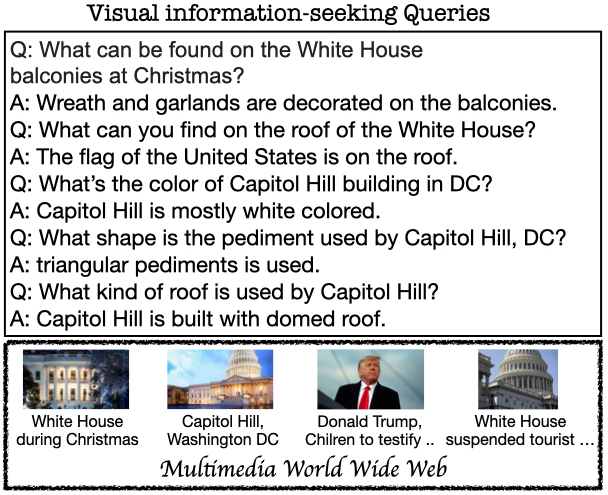}
    \caption{\textbf{Visual information-seeking queries}: These queries are unanswerable with text-only retrieval and require retrieving and reasoning over images.}
    \label{fig:intro_examples}
    \vspace{-3ex}
\end{figure}

Pre-trained language models like GPT-3~\cite{brown2020language}, PaLM~\cite{chowdhery2022palm}, etc have been shown to capture a massive amount of world knowledge implicitly in their parameters. However, using such large models incurs an extremely high computation cost. As an alternative to a singular monolithic transformer, retrieval-augmented architectures like KNN-LM~\cite{khandelwal2019generalization}, REALM~\cite{pmlr-v119-guu20a}, RAG~\cite{lewis2020retrieval}, FiD~\cite{izacard2021leveraging}, and RETRO~\cite{borgeaud2021improving} have been proposed to decouple world knowledge from the model's parameters. More specifically, these models are trained to access an external memory to enhance the model's predictions. Such retrieval-augmented architectures have multiple beneficial properties including: decreased model size~\cite{borgeaud2021improving}, better attribution/explanation for model predictions~\cite{lewis2020retrieval}, and adaptability to new information without retraining~\cite{verga2021adaptable}. However, previous retrieval-augmented models are limited to memories that contain only text or structured data and hence cannot make use of the massive amount of multimodal knowledge available on the web---much of which contains information only available in non-text modalities. 

~\autoref{fig:intro_examples}, shows several information-seeking queries that require retrieving and reasoning over visual knowledge. Here, a user first poses a question such as \textit{``What can be found on the White House balconies at Christmas''}. The system then retrieves relevant items from its memory, for example, the first image of ~\autoref{fig:intro_examples} with the caption \textit{``White House during Christmas''}, which it uses to produce the answer \textit{``wreaths and garlands''}. Existing text retrieval-augmented models would struggle with such queries because, in many cases, they would simply not have access to the answer as some knowledge does not exist in text form. That, coupled with the abundance of multimodal knowledge that exists, leads to the conclusion that retrieval-augmented models should ultimately be developed to retrieve and reason over multiple modalities.
\begin{figure}[!thb]
    \centering
    \includegraphics[width=0.98\linewidth]{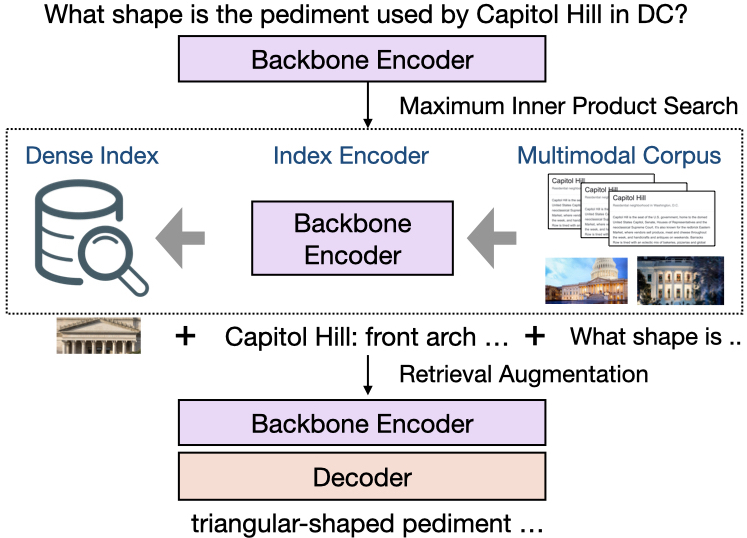}
    \caption{\textbf{Model Overview}: retrieval-and-predict process of \modelname on downstream datasets.}
    \label{fig:overview}
    \vspace{-2ex}
\end{figure}

In this paper, we are specifically interested in endowing pre-trained language models with a non-parametric multimodal memory containing images, text, or image-text pairs. To accomplish this, we first combine pre-trained T5~\cite{colin2020exploring} and ViT~\cite{dosovitskiy2020image} models to build a backbone encoder~(\autoref{fig:backbone}), which encodes image-text pairs, image-only, and text-only inputs into a multimodal representation. \modelname uses the backbone encoder to embed items into an external memory as well as queries to retrieve multimodal knowledge from that memory. These retrievals then augment a language model to generate more visually-grounded outputs.

We pre-train \modelname with a mixture of image-text and text-only datasets including LAION~\cite{schuhmann2021laion}, Conceptual-Caption~\cite{sharma2018conceptual}, VQA~\cite{antol2015vqa} and Probably-Asked-Questions (PAQ)~\cite{lewis2021paq}. More specifically, we reformulate these datasets in a retrieve-and-predict format. Here, the model's input is an image along with a text prompt. The model then retrieves from a memory containing captions and passages, which it uses to generate a target token sequence. The model is trained with both a contrastive and a generative loss; this teaches the model to discriminate relevant from irrelevant memory entries, and guides the model to leverage the multimodal knowledge into generation. 

Unlike the pre-training stage, during fine-tuning~\autoref{fig:overview} the model's input is a question, and the memory contains a collection of captioned images and text snippets. We fine-tune \modelname on the downstream datasets with a contrastive and generative loss similar to pre-training. To avoid excessive computation cost, we develop a two-stage training pipeline to first train with small in-batch memory, and then with a statically encoded and indexed large global memory. 

Our experiments show that \modelname achieves state-of-the-art performance on two different open-multimodal-QA datasets, both of which require retrieving images and text from a large corpus to answer factoid questions: WebQA~\cite{chang2021webqa} and MultimodalQA~\cite{talmor2021multimodalqa}. On both datasets, we outperform sophisticated baselines~\cite{li2020oscar,radford2021learning,zhang2021vinvl} by 10-20\% accuracy under both distractor (from 40+ candidates) and full-wiki settings (from 1M candidates). We also perform a comprehensive study to ablate different components of the pre-training to see their contributions. These empirical results demonstrate the effectiveness of our proposed models to integrate multimodal knowledge into pre-trained generation models and pave the way to unified retrieval-augmented frameworks.


\section{Related Work}
\noindent \textbf{Retrieval Augmented Models} Retrieval augmented models are hybrid models containing both parameterized sequence models and a non-parametric memory, infusing world knowledge into existing language models. Among them, KNN-LM~\cite{khandelwal2019generalization} was first proposed to retrieve instances from a text training corpus to help language modeling. Later, RETRO~\cite{borgeaud2021improving} was proposed to scale up the text corpus to trillions of tokens, enabling the model to achieve similar perplexity to GPT-3~\cite{brown2020language} with 25x fewer model parameters. Another family of models, such as REALM~\cite{pmlr-v119-guu20a}, RAG~\cite{lewis2020retrieval}, and FiD~\cite{izacard2021leveraging}, integrate Wikipedia passages as a datastore to benefit downstream knowledge intensive tasks (\textit{e.g.} Question Answering). REALM is an encoder-only model trained with masked language modeling, while RAG and FiD adopt an encoder-decoder model with a generative language modeling objective. Compared to them, \modelname is the first retrieval-augmented model that is capable of using knowledge presented in multiple modalities (\textit{i.e.} visual and textual knowledge data), whereas all prior methods are restricted to using text-only knowledge. \vspace{1ex} \\
\noindent \textbf{Multimodal Transformers} Multimodal transformers have demonstrated strong performances in learning cross-modal representation that are generally beneficial on downstream vision and language tasks, such as image-text retrieval~\cite{karpathy2015deep}, image captioning~\cite{chen2015microsoft}, and VQA~\cite{antol2015vqa}. These methods typically learn a joint transformer model on top of unimodal visual and textual backbones, via fusing deep features from each modality. The early version of multimodal transformers~\cite{lu2019vilbert,chen2020uniter,li2020oscar} usually learns a Transformer on pre-extracted unimodal features for contextualization, which makes it impossible to adjust those unimodal features to the target tasks. Recently, SimVLM~\cite{wang2021simvlm} and COCA~\cite{yu2022coca} proposed end-to-end training for both deep multimodal transformers and unimodal featurization networks and demonstrated strong performance in both multimodal and unimodal downstream tasks. The multimodal memory encoder of \modelname is broadly similar to SimVLM and CoCa, but has a different focus to encode and retrieve multimodal knowledge (\textit{i.e.} images and texts) to augment language generation models. \vspace{1ex} \\ 
\noindent \textbf{Multimodal Question Answering}
The problem of multimodal question answering has been extensively studied. VQA was the first proposed to answer questions from visual-only inputs. Later, OK-VQA~\cite{marino2019ok} enlarged VQA's scope to annotate questions requiring both image and implicit textual/common-sense knowledge to answer. More recently, MuMuQA~\cite{reddy2021mumuqa}, ManyModelQA~\cite{hannan2020manymodalqa} and MIMOQA~\cite{singh2021mimoqa} provide questions which require reasoning over images and explicitly provided text snippets. However, these datasets are restricted to dealing with given text and images without requiring any retrieval from the web: they are analogous to machine-reading approaches to QA from text like SQuAD, rather than open-book QA. To study the more realistic open multimodal QA task, WebQA~\cite{chang2021webqa} and MultimodalQA~\cite{talmor2021multimodalqa} have been proposed to evaluate answers to open queries which require retrieving and reasoning over a large-scale web multimodal corpus. Our model uses these datasets to study open-world multimodal question answering, obtaining state-of-the-art results.

\section{Model}
\subsection{Backbone Encoder}
\begin{figure}[!thb]
    \centering
    \includegraphics[width=1.0\linewidth]{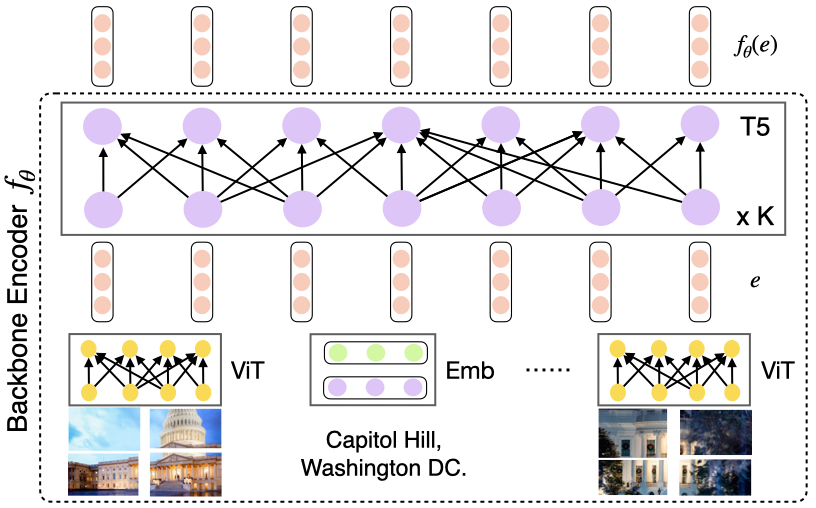}
    \caption{Backbone encoder: ViT encodes image patches into a sequence of vectors $\mathbf{e}_I$, while word embedding converts text tokens into another sequence of vectors $\mathbf{e}_T$. These vectors are concatenated to form $f_{\theta}(e)$ and fed to a decoder for text generation.}
    \label{fig:backbone}
    \vspace{-2ex}
\end{figure}
\modelname is built on top of a simpler model we call a ``backbone'' model, which is pre-trained to encode image-text pairs such that they are suitable for both answer generation and retrieval.  The backbone model's encoder is used as a component of the \modelname model. The backbone model is built with a pre-trained visual Transformer~\cite{dosovitskiy2020image} and a T5 text Transformer~\cite{colin2020exploring}, and consists of a multimodal encoder $f_{\theta}$ and decoder $g_{\theta}$. The encoder takes as input a sequence of image-text pairs, where either the image or the text component can be empty to accommodate text-only and image-only cases.

As depicted in~\autoref{fig:backbone}, the encoder can take a sequence of images and text. For image input, we first split each into 16x16 patches and feed them to a ViT~\cite{dosovitskiy2020image} transformer to generate a sequence of visual embedding denoted as $\mathbf{e}_I \in \mathbb{R} ^ {L_i \times D}$, where $L_i$ is the length of the image tokens. For text input, we use word embedding to produce another sequence of textual embedding $\mathbf{e}_T \in \mathbb{R} ^ {L_t \times D}$. For $k$ images and $n$ text inputs, we concatenate all their embeddings in the input order as $\mathbf{e}=[\mathbf{e}^1_I; \mathbf{e}_T^1; \cdots; \mathbf{e}^k_I; \mathbf{e}_T^n] \in \mathbb{R}^{(k L_t + nL_i) \times D}$, which is fed to another bi-directional transformer $f_{\theta}$ initialized from T5. We enable cross-attention between the two modalities to produce a fused representation, denoted as $f_{\theta}(\mathbf{e}) \in \mathbb{R}^{(kL_t + nL_i) \times D}$. We add a [CLS] token to obtain a pooled representation $f_{\theta}(\mathbf{e})_{\text{[CLS]}} \in \mathbb{R}^D$ for dense retrieval.

\subsection{\modelname}
\begin{figure*}[!thb]
    \centering
    \includegraphics[width=1.0\linewidth]{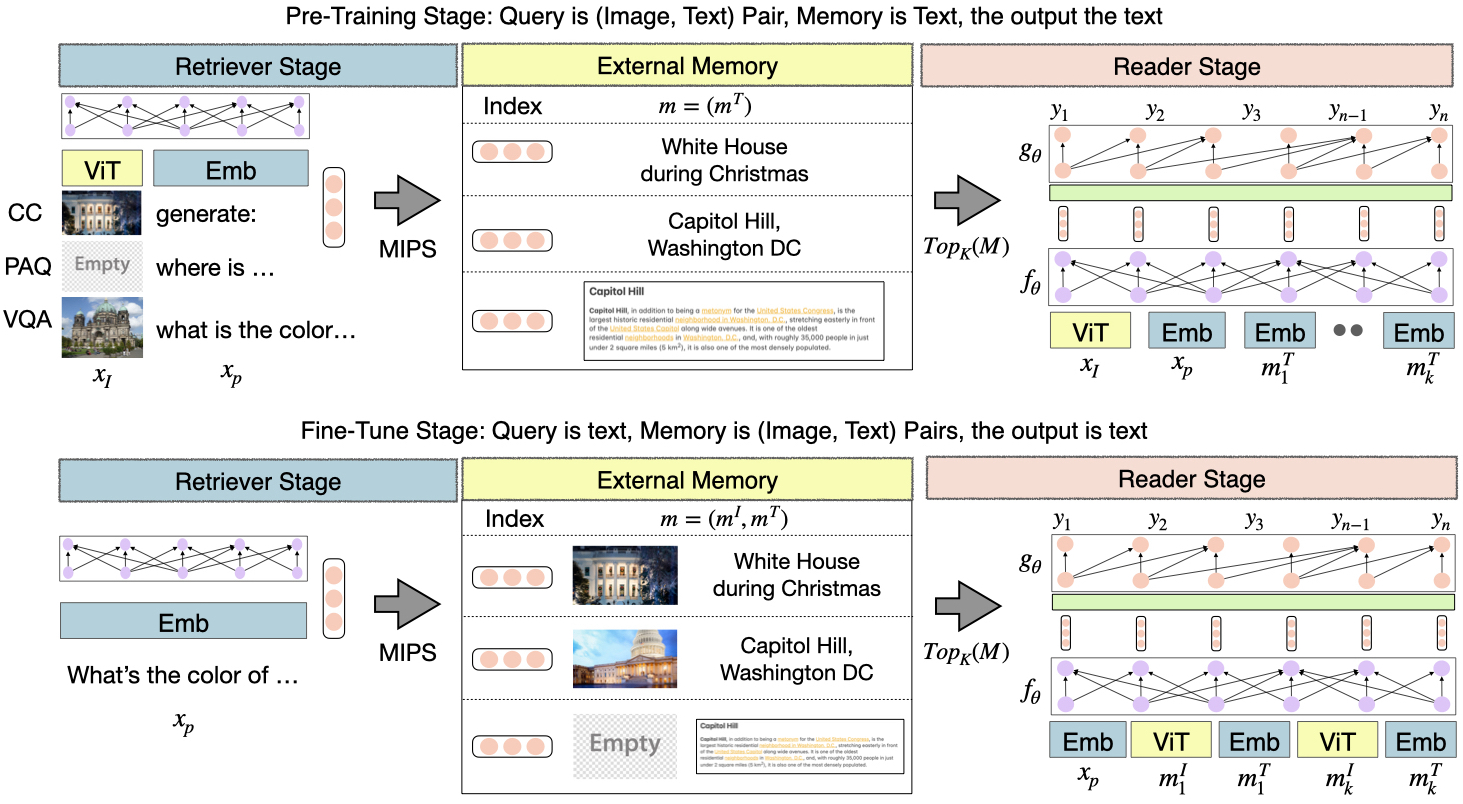}
    \caption{Model Architecture: the model accesses an external memory to obtain multimodal knowledge contained in images or text snippets, which is used to augment the generation. The upper part defines the pre-training implementation, while the lower part defines fine-tuning implementation.}
    \label{fig:murat}
    \vspace{-3ex}
\end{figure*}

We build \modelname (shown in ~\autoref{fig:murat}) on top of the backbone model. During the retriever stage, \modelname takes a query $q$ of any modality as input and retrieves from a memory $\mathcal{M}$ of image-text pairs. Specifically, we apply the backbone encoder $f_{\theta}$ to encode a query $q$, and use maximum inner product search (MIPS~\cite{guo2020accelerating}) over all of the memory candidates $m \in \mathcal{M}$ to find the Top-K nearest neighbors $Top_K(\mathcal{M}|q) = [m_1, \cdots, m_k]$. Formally, we define $Top_K(\mathcal{M}|q)$ as follows:
\begin{align*}
\begin{split}
    Top_K(\mathcal{M}|q) =  \underset{m \in \mathcal{\mathcal{M}}}{TopK} \quad   f_{\theta}(q)_{\text{[CLS]}} \cdot f_{\theta}(m)_{\text{[CLS]}}
\end{split}
\end{align*}
During the reader stage, the retrievals (the raw image patches) are combined with the query $q$ as an augmented input $[m_1, \cdots, m_k, q]$, which is fed to the backbone encoder $f_{\theta}$ to produce retrieval-augmented encoding. The decoder model $g_{\theta}$ uses attention over this representation to generate textual outputs $\mathbf{y} = y_1, \cdots, y_n$ token by token.
\begin{align*}
\begin{split}
    p(y_i|y_{i-1}) = g_{\theta}(y_{i} | f_{\theta}(Top_K(\mathcal{M}|q); q); y_{1:i-1})    
\end{split}
\end{align*}
where $y$ is decoded from a given vocabulary $\mathcal{V}$.


\subsection{Pre-training}
The pre-training implementation is depicted in the upper portion of~\autoref{fig:murat}, where the input query is an image $x_I$ plus a text prompt $x_p$. The external memory $\mathcal{M}$ contains textual-only entries $m^T$. The Top-K retrievals $m_1^T, \cdots, m_k^T$ are leveraged to generate the textual output. To avoid the excessive computation cost of backpropagation over the massive external memory, we adopt an in-batch memory $\mathcal{M}_B$, dynamically constructed from the input examples in a batch. The small in-batch memory enables \modelname to continuously update the memory encoder efficiently similar to TOME~\cite{de2021mention} and QAMAT~\cite{chen2022augmenting}.

\paragraph{Dataset}
The pre-training corpus consists of LAION~\cite{schuhmann2021laion}, Conceptual-Caption-12M+3M (CC)~\cite{sharma2018conceptual,changpinyo2021conceptual}, VQA~\cite{antol2015vqa} and PAQ~\cite{lewis2021paq}~\autoref{tab:pre-train}. LAION is a publicly-released image-text dataset containing crawled image-text pairs filtered by CLIP~\cite{radford2021learning}. We apply rules to filter LAION from 400M to 200M by removing text with HTTP-URLs or image width/height beyond 1000 pixels. CC contains 15M (image, anonymized alt-text) pairs crawled from the web but filtered more extensively to maintain high alignment quality. VQA contains annotated QA pairs aligned to MSCOCO images. We further add captions to each image from MSCOCO-Captioning~\cite{lin2014microsoft} to create (Image, Caption, QA) triples. PAQ is a text-only dataset containing 65M machine-generated QA pairs along with their source Wikipedia passage. 
\begin{table}[!htb]
\small
\centering
\begin{tabular}{lccc}
\toprule
Dataset        & \#Size &  Format           & Source    \\
\midrule
CC             & 15M    & (Image, Caption)   & Crawled   \\
LAION          & 200M   & (Image, Alt-Text)  & Crawled   \\
PAQ            & 65M    & (Passage, QA) & Generated \\
VQA            & 400K   & (Image, Caption, QA)  & Annotated \\
\bottomrule
\end{tabular}
\caption{Pre-training Dataset Statistics}
\vspace{-2ex}
\label{tab:pre-train}
\end{table}

For LAION and CC, we use the input image as $x_I$, and `generate caption:' as the text prompt $x_p$. For VQA, we use the input image as $x_I$ and the question as the prompt $x_p$. For PAQ, we use an empty array as the input image and the question as the prompt. The in-batch memory $\mathcal{M}_B$ is constructed by stacking the captions associated with the input images in LAION/CC/VQA and the passages associated with the questions in PAQ. Each textual memory entry is denoted as $m^T$. The decoder is optimized to generate either a caption or an answer, depending on the source dataset. Since the four dataset sizes are highly unbalanced, we use fixed mixture sampling ratios to balance their presence during pre-training. 

We train the model with a joint loss $L = L_{gen} + L_{con}$ as follows:
\begin{align*}
\small
\begin{split}
    L_{con} &= -\log \frac{exp(f_{\theta}(x_I, x_p) \cdot f_{\theta}(m^T))}{\sum_{m \in \mathcal{M}_B} exp(f_{\theta}(x_I, x_p) \cdot f_{\theta}(m^T))} \\
    L_{gen} &= -\log g_{\theta}(\mathbf{y} | f_{\theta}(M_p;x_I;x_p)) \\
    M_p &= \begin{cases}
    Top_K(\mathcal{M}_B|x_I, x_p) & \text{If } (x_I, x_p) \in \text{PAQ/VQA} \\
    $\O$ & \text{If } (x_I, x_p) \in \text{LAION/CC} \\
    \end{cases}
\end{split}
\end{align*}
where $M_p$ is the retrieved augmentation: if the input query is from PAQ/VQA, we use the retrieved memory entries, otherwise, we use null. The reason for setting it to null for LAION/CC is to avoid a trivial solution when the generation target (caption) also exactly appears in the memory. 

The contrastive loss $L_{con}$ is minimized to discriminate between the positive query-memory pairs and all other query-memory pairs from the memory. The pairwise matching score is computed as the dot product between query $f_{\theta}(x_I;x_p)_{\text{[CLS]}}$ and candidates $f_{\theta}(m^T)_{\text{[CLS]}}$. This objective enables the model to retrieve the most relevant knowledge from the memory. The generative loss $L_{gen}$ is minimized to generate target tokens $\mathbf{y}$ conditioned on the retrieval-augmented representation. This objective enables the model to combine information across different modalities for text generation.  


\subsection{Fine-tuning}
We finetune \modelname to align with the expected inputs of the downstream datasets which require answering text questions by retrieving image-caption pairs or text snippets from the external knowledge datastore. As depicted in the lower part of~\autoref{fig:murat}, the input query for the downstream task is a text question $x_q$, and the memory $\mathcal{M}$ containing (image, text) pairs $(m^I, m^T)$.\footnote{We set the image to a zero array if the memory entry is a text snippet.} The Top-K retrievals $\{(m_1^I, m_1^T), \cdots, (m_k^I, m_k^T)\}$ are leveraged to generate the answer $a$. To minimize the computation cost, we develop a two-stage pipeline to optimize with an in-batch memory and then resume with fixed retrieval from global memory.

\paragraph{In-Batch Training}
In this stage, we aim to minimize the joint loss function $L = L_{con} + L_{gen}$ based on the in-batch memory $\mathcal{M}_B$ as follows:
\begin{align*}
\small
\begin{split}
    L_{con} &= -\log \frac{exp(f_{\theta}(x_q) \cdot f_{\theta}(m^I;m^T))}{\sum_{m \in \mathcal{M}_B} exp(f_{\theta}(x_q) \cdot f_{\theta}(m^I;m^T))} \\
    L_{gen} &= -\log g_{\theta}(\mathbf{y} | f_{\theta}(Top_K(\mathcal{M}_B|x_q) ;x_q))
\end{split}
\end{align*}
The in-batch memory $\mathcal{M}_B$ is constructed in the following way: the $k$-th example in the dataset is represented as $(x_{q,k}, y_k, \{m^I_i, m^I_i\}_k, \{\bar{m}^I_j, \bar{m}^T_j\}_k)$, where $m$ represents the positive (image, text) source, and $\bar{m}$ represents the hard negative (image, text) source provided by the dataset\footnote{These hard negatives are mined through Bing Search API and Wikipedia page, refer to~\cite{chang2021webqa} for details.}. For a batch with $B$ examples, we assemble all the associated positive and negative knowledge source as our in-batch memory $\mathcal{M}_B=\{\{m^I_i, m^I_i\}_1, \{\bar{m}^I_j, \bar{m}^T_j\}_1, \cdots, \{\bar{m}^I_j, \bar{m}^T_j\}_B\}$.


\paragraph{Fixed-Retrieval Training}
After in-batch training, we encode all available cross-modal pairs, and index these encodings for fast MIPS retrieval.  We then apply the trained retriever to search over the full multimodal corpus $\mathcal{M}$ to obtain the global top-K retrievals $Top_K(\mathcal{M}|x_q)$ and continue to optimize $L_{gen}$.  During this training phase, the stored encodings are not updated.
\nop{
\begin{align*}
\small
\begin{split}
    L_{gen} &= g_{\theta}(\mathbf{y} | f_{\theta}(Top_K(\mathcal{M}|x_q);x_q))
\end{split}
\end{align*}
}
During inference time, we use fixed encodings to generate the answers.

\section{Experiments}
\subsection{Implementation Details}
The backbone model uses T5-base~\cite{colin2020exploring} and a ViT-large model~\cite{dosovitskiy2020image} as described in~\autoref{tab:model_size}. We adopt the sentence-piece model from T5 with a vocabulary size of 32128. The ViT model was pre-trained on the JFT dataset. We resize every image into 224x224 pixels and split them into a sequence of 16x16 patches. The output of ViT is a sequence of 1024-dimension vectors, which are projected to 768-dimension for consistency with T5 model. \modelname reuses the model as retriever and reader, thus the full model size is 527M parameters. 
\begin{table}[!thb]
\centering
\small
\begin{tabular}{cccccc}
\toprule
Model & \#Enc & \#Dec & Hidden & Heads & Params \\
\midrule
ViT-large &  24 & 0 & 1024  & 16 & 307M \\
T5-base  & 12 & 12 & 768 & 12 & 220M \\
\bottomrule
\end{tabular}
\caption{The model size and configurations, with \#Enc/\#Dec denoting encoder/decoder layers.}
\label{tab:model_size}
\vspace{-2ex}
\end{table}

Our model is implemented in JAX~\cite{jax2018github}, based on the T5X codebase~\cite{roberts2022scaling}. During pre-training, we first train the model on LAION for 1M steps, and then continue training on CC/PAQ/VQA with 1:1:1 sample ratio for another 200K steps. We optimize the model with Adafactor~\cite{shazeer2018adafactor}. For both stages, we adopt a constant learning rate of 5e-4 and a batch size of 4096. The models are trained on 64 Cloud v4 TPUs~\cite{jouppi2020domain}.

We then fine-tune \modelname on WebQA and MultimodalQA with a constant learning rate of 3e-4 for 20K steps. The checkpoint with the highest validation score is run on the test set. We use a batch size of 64 and set TopK=4 for both in-batch training and fixed-retrieval training. We noticed that increasing Top-K further does not yield further improvement. We use a beam size of 2 to search for the best hypothesis for both datasets (increasing it further doesn't yield better performance). 

\subsection{Datasets}
For evaluation, we choose two multimodal QA datasets: WebQA~\cite{chang2021webqa} and MultimodalQA~\cite{talmor2021multimodalqa} and demonstrate their statistics in~\autoref{tab:finetune-data}.
\begin{table}[!htb]
\centering
\small
\begin{tabular}{lccc}
\toprule
Dataset      & Train     & Dev       & Test      \\
       & Image/Text & Image/Text & Image/Text \\
\midrule
WebQA        & 18K/17K   & 2.5K/2.4K & 3.4K/4K   \\
MultimodalQA & 2.1K/7.4K & 230/721   & -       \\
\bottomrule
\end{tabular}
\caption{Overall Statistics of downstream dataset.}
\label{tab:finetune-data}
\vspace{-3ex}
\end{table}

\paragraph{WebQA} 
This dataset contains multi-hop, multimodal question-answer pairs where all questions are knowledge-seeking queries. The queries require 1-2 images or 1-2 text snippets to answer. Each query in WebQA is associated with a set of visual/text distractors (hard negatives). The answers in WebQA are normally complete sentences to better assess the model's generation capability. Two evaluation setups are used, namely distractor and full-wiki. Under the distractor setup, the model needs to retrieve from these hard negatives + positives to answer the question. Under the full-wiki setup, the model needs to search over 1.1M text and visual sources from Wikipedia to answer the question. For evaluation, WebQA uses  BARTScore~\cite{yuan2021bartscore} to measure the fluency between the generation and the reference, and keyword accuracy score to measure the correctness/truthfulness of the generation. These two scores are multiplied to calculate the overall score.

\paragraph{MultimodalQA-Subset}
This dataset contains human-annotated multimodal questions over different modalities including tables, text, and images. Wikipedia tables are used as anchors to connect different modalities. The authors first use the template to generate questions and then ask crowd-workers to filter and paraphrase the generated questions. Since tables are outside the scope of our paper, we focus on the subset of queries requiring only text and image information. Specifically, we choose the questions with types of `TextQ' and `ImageQ' to construct the subset. The query requires 1 image or 1 text snippet to answer. Each query in MultimodalQA is also associated with visual and text distractors (hard negatives). Similarly, two evaluation setups are used as before. Under a full-wiki setup, MultimodalQA uses a database containing 500K text and visual sources. The evaluation scores are based on Exact Match and F1.

\subsection{Baselines}
For WebQA and MultimodalQA, we mainly compare different variants of pre-trained vision-language models. 
\paragraph{VLP}
In WebQA, VLP-like models~\cite{zhou2020unified} like Oscar~\cite{li2020oscar} and VinvL~\cite{zhang2021vinvl} are used as the standard baselines. These models were pre-trained on Conceptual 3M~\cite{sharma2018conceptual} with a masked language objective. During fine-tuning, the VLP model takes a set of token inputs <[CLS], $s_i$, [SEP], Q, [SEP]> first to select the most plausible source $s_i$, and then feed $s_i$ in the form of <[CLS], $S$, $Q$, $A$, [SEP]> to autoregressively decode answer $A$ with masked language model prediction.
\paragraph{AutoRouting}
In MultimodalQA, this method first applies a question type classifier to detect the modality of the question (either a passage or an image), and then routes the question to its sub-model. The method uses RoBERTa-large~\cite{roberts2022scaling} for text-questions and VilBERT~\cite{lu2019vilbert} with features extracted from Faster-RCNN~\cite{ren2015faster} for image questions.
\paragraph{CLIP (K)}
CLIP~\cite{radford2021learning} is used for full-wiki retrieval. Specifically, the baselines systems adopt CLIP to encode queries and all the image/text candidates separately into vectors and then run approximated nearest neighbor searches to find a set of K potential candidates. After the coarse-level retrieval without cross-attention, it adopts a reranker to further narrow down to the 1-2 candidates to feed as input $S$ to the QA model.

\subsection{Experimental Results}
We demonstrate WebQA's results in~\autoref{tab:webqa}. All results reported are the medium score from three runs with different random seeds, and the variance of the Overall score is within 0.2\%. We can observe that \modelname can significantly outperform VLP with different backends including Oscar, ResNet, and VinVL. In retrieval performance, our model outperforms VLP by 15\% in the full-wiki setting. For Fluency, our model outperforms VLP by 12\% under the distractor setting and 14\% under the full-wiki setting. For Accuracy, our model manages to achieve 16\% under the distractor setting and even 20\% the under the full-wiki setting. These improvements reflect the high fluency and accuracy of \modelname's generation, and the improvement is more pronounced for full wiki.

\begin{table}[!tb]
\small
\centering
\begin{tabular}{l|cccc}
\toprule
Evaluation       & \multicolumn{4}{c}{Distractor}                                                                                                     \\
Metrics          & Retr & FL                      & Accuracy                         & Overall                      \\ 
\midrule
Question-Only     & -      & 34.9                           & 22.2                          & 13.4 \\ 
VLP (Oscar)       & 68.9   & 42.6                           & 36.7                           & 22.6                                                    \\
VLP + ResNeXt    & 69.0       & 43.0                           & 37.0                           & 23.0                                                  \\
VLP + VinVL       & 70.9    & 44.2                           & 38.9                           & 24.1                                                    \\
\midrule
\modelname     & \textbf{74.6}        & \textbf{55.7} & \textbf{54.6} & \textbf{36.1}   \\
\midrule
\midrule
Evaluation      & \multicolumn{4}{c}{Full-Wiki}                                                                                                       \\
\midrule
CLIP (2) + VLP  & 11.9   & 34.2                           & 24.1                           & 14.6                                                      \\
CLIP (20) + VLP & 24.0   & 36.1                           & 27.2                           & 16.1                                                     \\ \midrule
\modelname           & \textbf{39.7}  & \textbf{50.7} & \textbf{47.8} & \textbf{31.5}   \\
\bottomrule
\end{tabular}
\caption{WebQA official test-set results indicated on leaderboard\footnote{\url{https://eval.ai/challenge/1255/leaderboard/3168}} as of May 2022. Retr denotes the retrieval-F1 score. FL refers to fluency metric BARTSCcore, and Accuracy refers to keyword matching F1 score, they are combined as Overall. }
\vspace{-3ex}
\label{tab:webqa}
\end{table}

We show the MultimodalQA results in~\autoref{tab:mmqa}. We can see that \modelname is also able to vastly outperform the routing-based multimodality QA model. For text questions, our model improves over AutoRouting by 10+\% EM under both settings. For image questions, the gap becomes more significant, with 20+\% improvement under both settings. Similarly, we find that our model is more capable of handling full-wiki corpus. 
\begin{table}[!tb]
\centering
\small
\begin{tabular}{lccccc}
\toprule
Evaluation                         & \multicolumn{4}{c}{Distractor}                                            \\
\multirow{2}{*}{Metrics} & \multicolumn{2}{c}{Text}        & \multicolumn{2}{c}{Image}       & All  \\
                         & EM   & \multicolumn{1}{c}{F1}   & EM   & \multicolumn{1}{c}{F1}   & EM   \\ 
\midrule
Question-Only            & 15.4 & 18.4 & 11.0 & 15.6 & 13.8 \\
AutoRouting              & 49.5 & 56.9 & 37.8 & 37.8 & 46.6 \\
\midrule
\modelname                   & \textbf{60.8}  & \textbf{67.5}   & \textbf{58.2}  & \textbf{58.2}  & \textbf{60.2} \\ 
\bottomrule
\end{tabular}
\begin{tabular}{lccccc}
\toprule
Evaluation                           & \multicolumn{4}{c}{Full-Wiki}                                             \\
\multirow{2}{*}{Metrics} & \multicolumn{2}{c}{Text}        & \multicolumn{2}{c}{Image}       & All  \\
                         & EM   & \multicolumn{1}{c}{F1}   & EM   & \multicolumn{1}{c}{F1}   & EM   \\ 
\midrule
\begin{tabular}[c]{@{}l@{}}CLIP (10) + \\ AutoRouting\end{tabular}           & 35.6  &  40.2   & 32.5  &  32.5  &  34.7  \\
\midrule
\modelname                     &   \textbf{49.7}   & \textbf{56.1}  &  \textbf{56.5}    &  \textbf{56.5}  &  \textbf{51.4} \\
\bottomrule
\end{tabular}
\caption{Multimodal dev-set results on the subset.}
\vspace{-1ex}
\label{tab:mmqa}
\end{table}

\subsection{Ablation Study}
Here we ablate the properties of \modelname to better understand our experimental results.
\paragraph{Pre-training Corpus}
In order to study the contributions of different pre-training corpora, we investigated several pre-training corpus combinations.  We report their fine-tuned results on WebQA test set in~\autoref{tab:pre-train-ablation}. As can be seen, without any pre-training, our model only achieves an overall score of 23.5, which lags behind the baseline models. After pre-training on different singular datasets, \modelname is able to achieve better performance than the baselines. Among the individual datasets, LAION is shown to yield the highest score, and adding CC, PAQ, and VQA to the pre-training corpus set one by one produces steady improvements. 
\begin{table}[!tb]
\small
\centering
\begin{tabular}{lccc}
\toprule
Pre-train Dataset & FL   & Accuracy & Overall \\
\midrule
None              & 42.5 & 36.1     & 23.5    \\
\midrule
CC                & 46.4 & 41.3     & 25.6    \\
LAION             & 47.8 & 44.8     & 28.3    \\ 
VQA               & 47.0 & 44.4     & 27.4    \\ 
PAQ               & 46.8 & 42.8     & 27.0    \\
\midrule
LAION+CC          & 49.5 & 47.4     & 30.7    \\
LAION+CC+PAQ      & 53.7 & 51.8     & 34.4   \\
LAION+CC+PAQ+VQA  & 55.7 & 54.6     & 36.1   \\
\bottomrule
\end{tabular}
\caption{Ablation Study for different pre-training corpus, score under distractor setting.}
\label{tab:pre-train-ablation}
\vspace{-2ex}
\end{table}

\paragraph{Two-Stage Fine-tuning}
In order to study the necessity of the two-stage fine-tuning, we perform an ablation study to see the impact of the two stages. We display our results in~\autoref{tab:two-stage-ablation}. (Only In-Batch) refers to the model trained only with in-batch memory are directly used to generate outputs by accessing the global memory. Without further tuning, the performance will drop by roughly 2\% on both datasets. (Only Fixed-Retrieval) refers to using the pre-trained retriever directly to obtain Top-K and then optimize the generative loss. As can be seen, the performance drop is more severe in this case for both datasets. This is understandable due the misalignment between pre-training retrieval is (image + text->text) while the fine-tuning retrieval is (text -> image+text).  Thus, it is necessary to adapt the \modelname's pre-trained retriever to different use cases depending on the downstream datasets. 

\begin{table}[!tb]
\small
\centering
\begin{tabular}{lcc}
\toprule
Model &  WebQA & Multimodal \\
\midrule
\modelname (Only In-Batch)              & 29.4   &  49.6 \\
\modelname (Only Fixed-Retrieval)       & 25.8   & 40.7 \\
\modelname (Two Stage)       & 31.5   &  51.4 \\
\bottomrule
\end{tabular}
\caption{Ablation Study for different fine-tuning stages to see their contributions. WebQA uses the overall score, and MultimodalQA refers to EM-all score.}
\label{tab:two-stage-ablation}
\vspace{-1ex}
\end{table}

\subsection{Human Analysis}
In order to better understand the model's performance, we manually study 200 model outputs and classify them into three categories and show our manual analysis results in~\autoref{tab:error-analysis}. As can be seen, image queries are much harder than text queries. \modelname only achieves 64\% accuracy for the distractor setting and 54\% accuracy for the full-wiki setting, falling significantly behind text accuracy. 

\begin{figure}[!thb]
    \centering
    \includegraphics[width=0.95\linewidth]{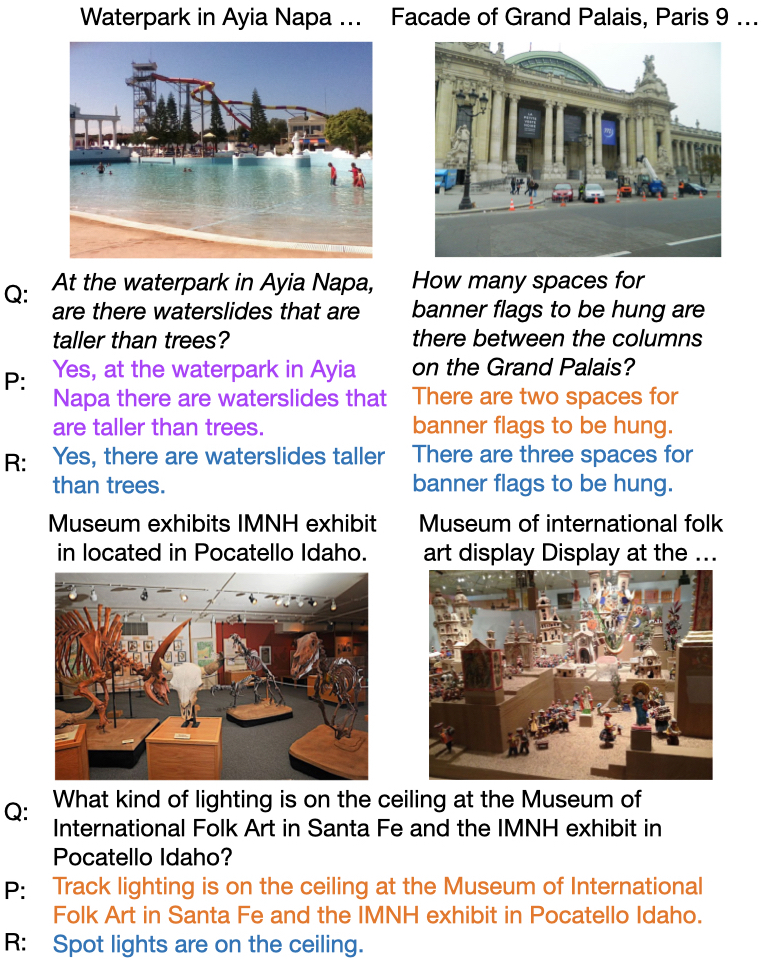}
    \caption{Upper left: correct prediction, Upper Right: error due to miscounting, Lower: error due to misrecognition (multiple image reasoning). Q refers to the question, P refers to prediction and R refers to the reference.}
    \label{fig:examples}
    \vspace{-2ex}
\end{figure}

\begin{table}[!tb]
\small
\centering
\begin{tabular}{llcc}
\toprule
Evaluation                  & Model     & Correct & Wrong\\ 
\midrule 
\multirow{2}{*}{Distractor} & \modelname (Text)  & 80\%    & 20\%      \\
                            & \modelname (Image) & 64\%    & 36\%        \\ 
\midrule 
\multirow{2}{*}{Full-Wiki}  & \modelname (Text)  & 72\%    & 28\%        \\
                            & \modelname (Image) & 54\%    & 46\%      \\ 
\bottomrule
\end{tabular}
\vspace{-2ex}
\caption{The human evaluation results on WebQA dataset separately for image/text queries.}
\label{tab:error-analysis}
\vspace{-2ex}
\end{table}

We further categorize the image-query errors manually into the categories of~\autoref{tab:error-category}. Counting is the most difficult question type, and constitutes 52\% of the total errors, while object recognition errors rank second, constituting 29\% of errors. In contrast, identifying color, shape, and gender is comparatively easier, with fairly low error rates. We demonstrate some correct and typical error cases in~\autoref{fig:examples} including miscounting and misrecognizing objects. We observe that these errors are mostly due to several reasons: 1) the question is related to infrequent objects, thus making recognition errors, 2) the image scene is highly complex with a large number of objects, thus grounding to a specific region is difficult, 3) the questions require optical character recognition ability from images. Hence, the bottleneck of \modelname is still in the visual understanding module. 

\begin{table}[!thb]
\small
\centering
\begin{tabular}{lccccc}
\toprule
Category & Count &  Object & Color & Shape & Gender \\
\midrule
Ratio  &  52\%    &   29.4\%   &  5.8\%     &  5.8\% & 5.8\%  \\
\bottomrule
\end{tabular}
\caption{Error categorization and their ratios on sampled WebQA-dev image queries. }
\label{tab:error-category}
\vspace{-2ex}
\end{table}

\section{Examples}
\begin{figure}[!thb]
    \centering
    \includegraphics[width=0.98\linewidth]{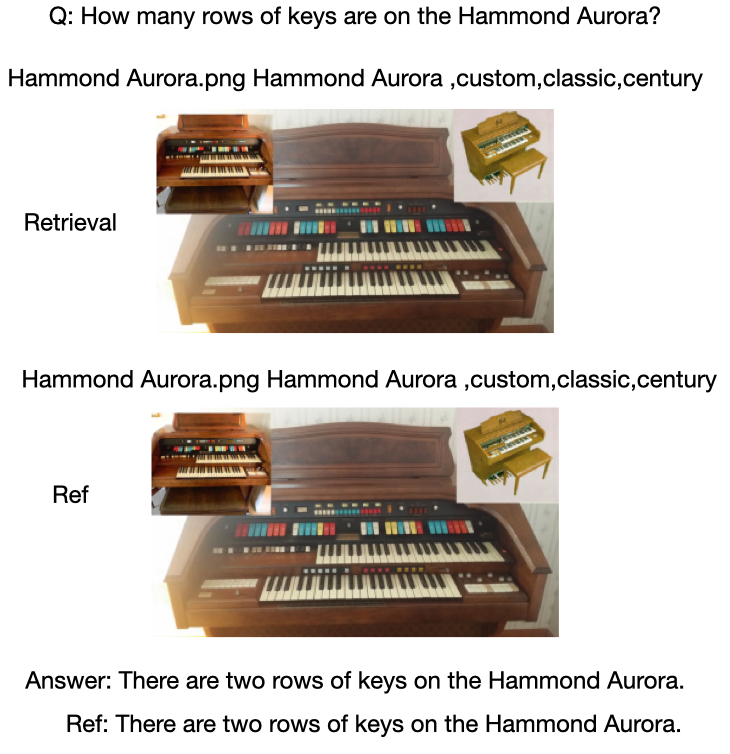}
    \caption{Examples: we demonstrate model retrieval vs. groundtruth and model answer vs. reference. }
    \label{fig:examples1}
\end{figure}

\begin{figure}[!thb]
    \centering
    \includegraphics[width=0.98\linewidth]{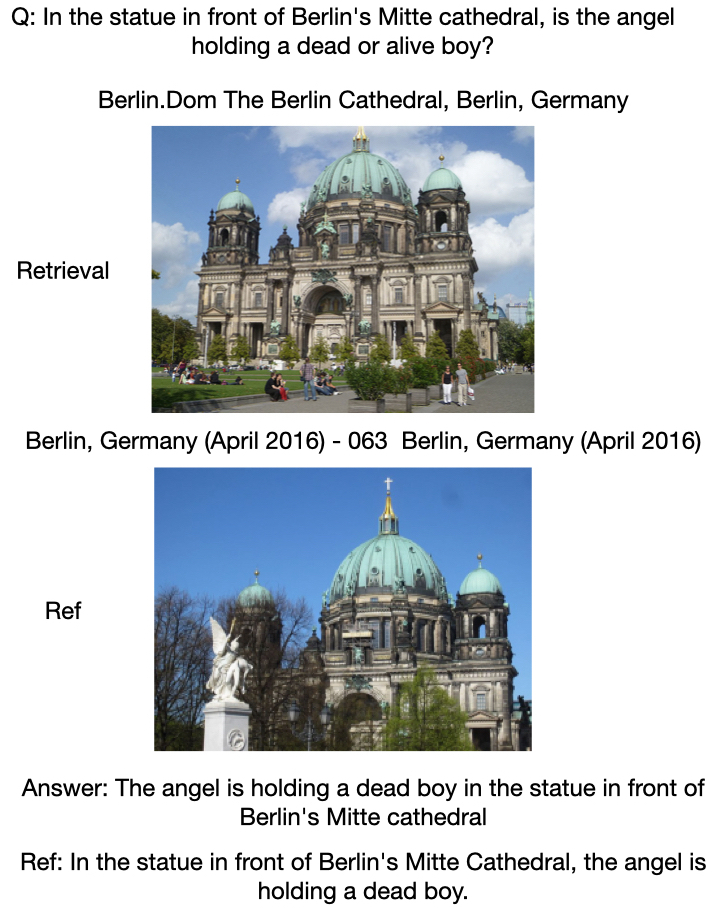}
    \caption{Examples: we demonstrate model retrieval vs. groundtruth, and model answer vs. reference. }
    \label{fig:examples2}
\end{figure}

We list more examples in~\autoref{fig:examples1} and~\autoref{fig:examples2}. As can be seen, in the first example, the model is grounded on the oracle image-text pair to make the correct prediction. However, in the second example, though the model retrieves the wrong image-text pair, it is able to make the correct prediction of `the angel is holding a dead body'. We conjecture that the model utilizes textual clues to make the prediction rather than grounding on the image itself. Such shortcut learning is concerning and needs to be addressed through better learning algorithms.

\section{Conclusion}
In this paper, we build the first visually-grounded language generator capable of retrieving multimodal knowledge from a large-scale corpus. Our experiments show the promise of this approach, as it outperforms existing baselines by a large margin. At the same time, the performance on knowledge-seeking queries that require reasoning over images is still significantly lower than the performance on queries requiring only text. This indicates that there is still ample room for further improvements and we hope our study can motivate more research on better multimodal retrieval-augmented models.

\section*{Limitations}
The current approach has several limitations: 1) since we do not mine hard negatives during pre-training, negatives come from other examples within the same batch. This requires that we set the batch size sufficiently large enough to collect hard-enough negatives. This results in the pre-training requiring a large number of computation resources to reach competitive retrieval abilities. 2) our pre-training corpus's format (image -> text) is different from fine-tuning (text -> image+text). This misalignment limits the model's performance. Future work should consider how to design a better-aligned pre-training objective to achieve better performance. 3) Current visual representation in the reader stage is relatively expensive, i.e. 16x16=196 tokens per image, which poses great challenges for the transformer encoder to scale up to large Top-K values due to the quadratic attention complexity.

\section*{Ethical Statement}
Our work uses the LAION dataset, a widely-used and publicly available large-scale visual-language corpus crawled from the web. The authors have conducted automatic filtering to greatly reduce harmful content. However, it is not possible to fully remove all of the potential risks from the data given its tremendous size. Being trained on this dataset, we anticipate our model to contain some biases (racial, gender, etc.). During our manual inspection, we saw some such biases, for example, 5\% of errors are caused by misrecognition of gender. However, there are other many other forms of biases that we cannot fully enumerate or observe explicitly.

\bibliography{anthology,custom}
\bibliographystyle{acl_natbib}
\clearpage

\appendix

\section{Pre-training}
\begin{figure}[!thb]
    \centering
    \includegraphics[width=1.0\linewidth]{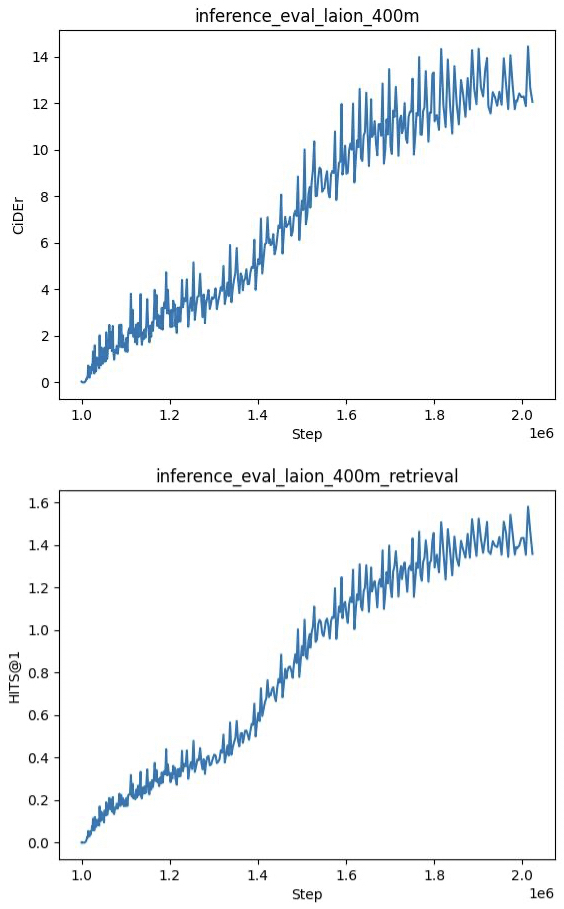}
    \caption{LAION Pre-training, validation accuracy, generation Cider score and retrieval recall score from the in-batch memory. }
    \label{fig:laion}
    \vspace{-2ex}
\end{figure}

During Pre-trainnig, we found that directly training with a mixture of all four datasets will lead to instability. We experimented with different variants and found that a scheduled pre-training can lead to a stable solution. We propose to first pre-train the model on the largest LAION dataset for 1M steps, and then continue training on the other three datasets with a fixed sample ratio. We plot the first stage of LAION training in~\autoref{fig:laion}. We monitor the generation quality (LAION image -> text captioning), and the retrieval quality (image -> 4096 in-batch caption retrieval). As can be seen, the LAION pre-training converges after 1M steps, where we first warm up and then decrease the learning rate using a scheduler.

\begin{figure}[!thb]
    \centering
    \includegraphics[width=0.92\linewidth]{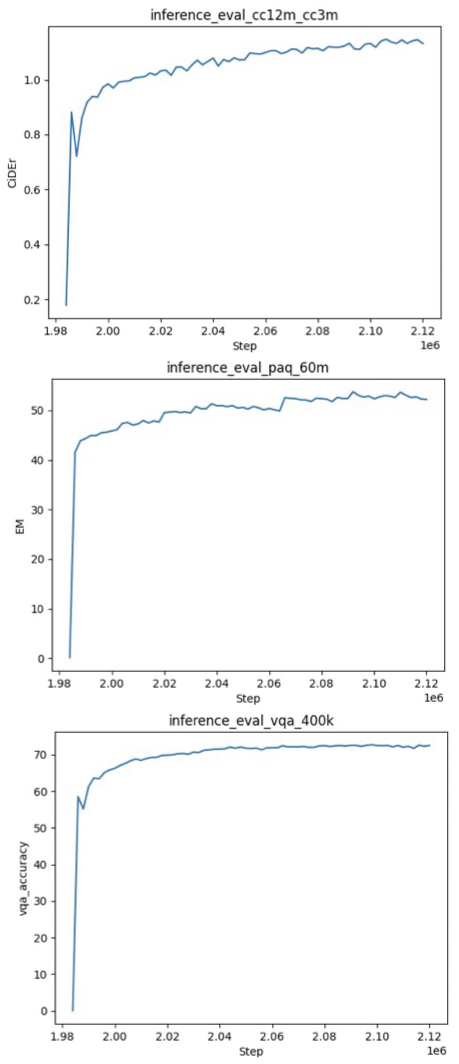}
    \caption{Mixture Pre-training, CiDEr, EM, and VQA accuracy for CC, PAQ, and VQA datasets. }
    \label{fig:mixture}
    \includegraphics[width=0.9\linewidth]{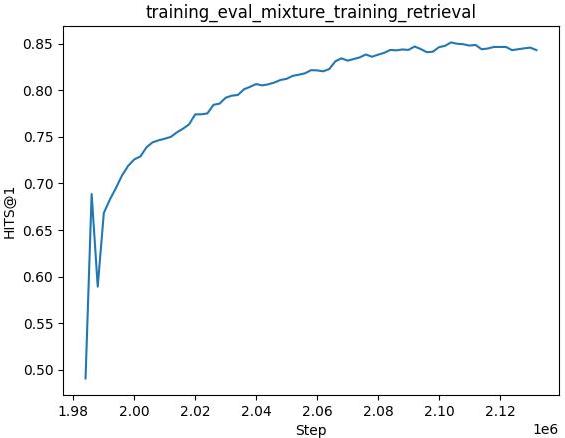}
    \caption{Mixture Pre-training retrieval accuracy over CC, PAQ, and VQA datasets. }
    \label{fig:mixture_retrieval}
    \vspace{-2ex}
\end{figure}

We further the pre-training on a mixture of the other three datasets. We plot their inference evaluation scores in~\autoref{fig:mixture}. We can see that the model is able to achieve very strong performance on these datasets, i.e. higher than 1.2 CiDEr on CC12M+3M validation set. The model also achieves strong performance on text-only reading comprehension on PAQ (similar to NQ), i.e. higher than 55\% EM score. On the VQA dataset, the model is able to achieve higher than 72\% VQA accuracy on the validation set. These results demonstrate the efficiency and multi-tasking capabilities of the pre-trained model.  The overall retrieval accuracy from the multimodal memory consisting of captions, and passages are plotted in~\autoref{fig:mixture_retrieval}, where the model is able to achieve 85\% RECALL@1 from a 4K memory. 

\section{Model Configuration}
We demonstrate the ViT configuration as follows:
\begin{lstlisting}
"vit_config": {
  "model": "ViT",
  "patches": {
    "size": [16, 16]
  },
  "hidden_size": 1024,
  "image_size": [224, 224],
  "num_heads": 16,
  "num_layers": 24,
  "mlp_dim": 4096,
  "return_pooled_output": false,
  "dropout_rate": 0.1
},
\end{lstlisting}
We demonstrate the T5-EncDec configuration as follows:
\begin{lstlisting}
"model_config": {
  "vocab_size": 32128,
  "hidden_size": 768,
  "intermediate_dim": 2048,
  "num_attention_heads": 12,
  "memory_key_dim": 768,
  "encoder_layers": 12,
  "decoder_layers": 12,
  "dropout_rate": 0.1,
  "max_distance": 128,
  "num_buckets": 32,
  "scale": 1.0,
  "retrieval_weight": 0.5,
}
\end{lstlisting}

\end{document}